\documentclass{article}

\usepackage{PRIMEarxiv}

\usepackage[utf8]{inputenc} 
\usepackage[T1]{fontenc}    
\usepackage{hyperref}       
\usepackage{url}            
\usepackage{booktabs}       
\usepackage{amsfonts}       
\usepackage{nicefrac}       
\usepackage{microtype}      
\usepackage{fancyhdr} 
\usepackage{graphicx}
\graphicspath{{media/}}     

\graphicspath{{figures/}}
\usepackage{natbib}

\usepackage{siunitx}
\DeclareSIUnit\angstrom{\text {Å}}
\usepackage{tabularx}
\usepackage{makecell}

\usepackage{amsmath}

\usepackage{subcaption}
\usepackage{comment}
\title{Aligning Text-to-Image Diffusion Models with Noise-Conditioned Perception
\thanks{\textit{\underline{Citation}}:
\textbf{A. Gambashidze, A. Kulikov, Y. Sosnin and I. Makarov, "Aligning Text-to-Image Diffusion Models With Noise-Conditioned Perception," in IEEE Access, vol. 13, pp. 193745-193753, 2025, doi: 10.1109/ACCESS.2025.3632092.}}
}

\author{
  Alexander Gambashidze \\
  Artificial Intelligence Research Institute (AIRI) \\
  Skolkovo Institute of Science and Technology \\
  Moscow \\
  \texttt{gambashidze@airi.net} \\
  \And
  Anton Kulikov \\
  HSE University \\
  Moscow \\
  \And
  Yuriy Sosnin \\
  HSE University \\
  Moscow \\
  \And
  Ilya Makarov \\
  Artificial Intelligence Research Institute (AIRI) \\
  Research Center for Trusted Artificial Intelligence, ISP RAS \\
  Moscow \\
}

\begin{document}
\maketitle

\begin{abstract}
Human preference optimization, originally developed for Language Models, has shown promise in improving text-to-image Diffusion Models by enhancing prompt alignment, visual appeal, and user preference. However, Diffusion Models are typically optimized in pixel or VAE space, which often misaligns with human perception, resulting in slower and less efficient training during fine-tuning and preference optimization. In this work, we demonstrate that using a perceptual objective significantly enhances both training speed and overall model quality. We fine-tune Stable Diffusion 1.5 and XL using Direct Preference Optimization (DPO), Contrastive Preference Optimization (CPO), and supervised fine-tuning (SFT) within this perceptual embedding space. Our approach significantly outperforms standard latent-space implementations across various metrics, including quality and computational cost, when training on the widely-used Pick-a-Pic dataset. For SDXL, our method achieves a 64.6\% general preference over the baseline DPO on the PartiPrompts dataset while significantly reducing compute to reach comparable DPO performance. Additionally, we enhance the Pick-a-Pic dataset, making it approximately 10 times smaller while training models that surpass the original published versions in just 12 and 3.5 GPU hours for SDXL and SD1.5, respectively. This paper demonstrates that the overall quality of Diffusion models after fine-tuning can be significantly improved while also being more efficient and requiring far less data.
\end{abstract}


\begin{figure*}[htbp]
    \centering
    \includegraphics[width=1\linewidth]{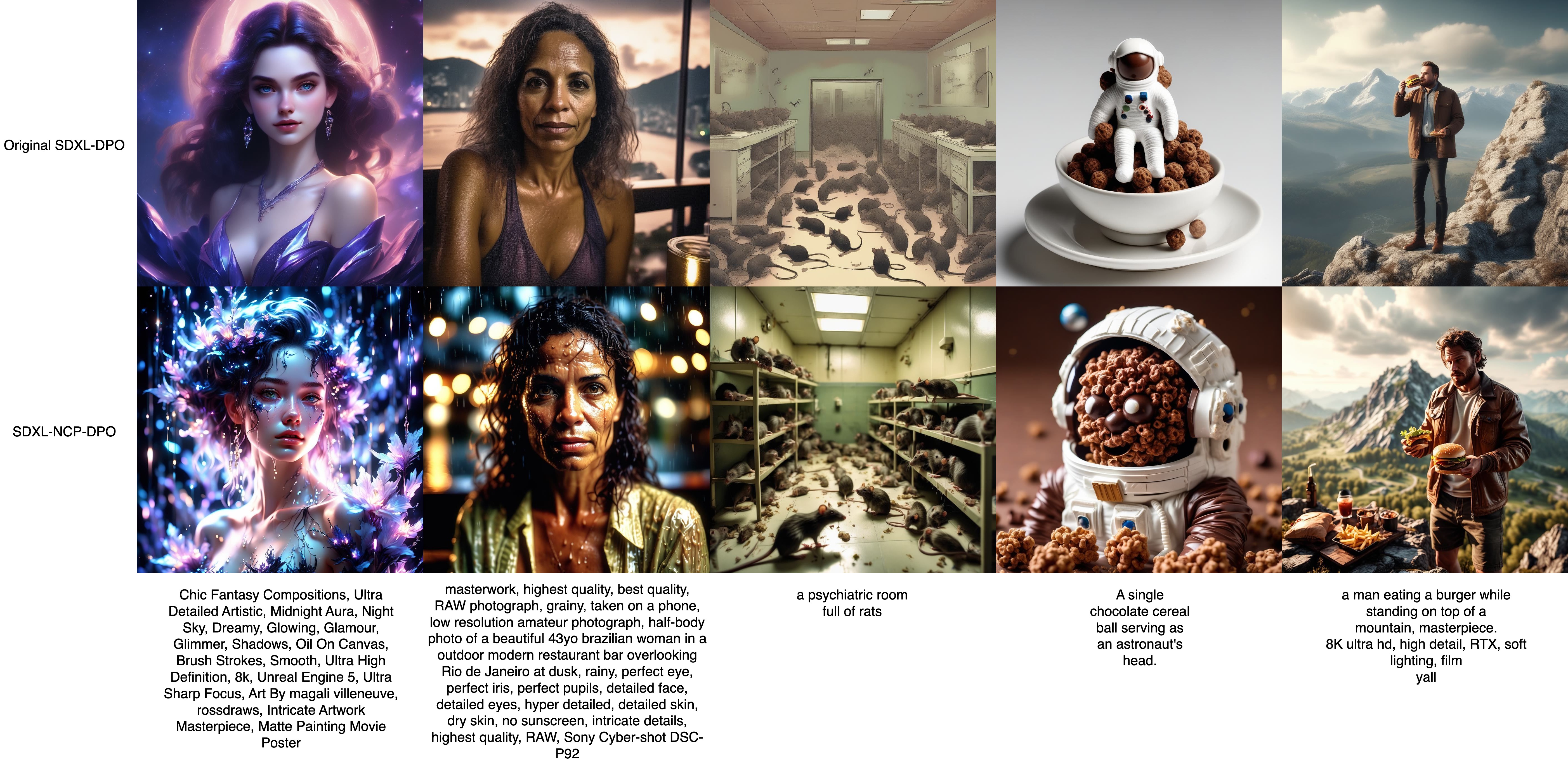}
    \caption{Noise-Conditioned Perceptual objective for aligning diffusion models significantly improves Direct Preference Optimization. }
    \label{fig:enter-label}
\end{figure*}

\section{Introduction}

Recent research has highlighted the benefits of alignment—the process of fine-tuning a generative model to meet specific objectives, particularly human preferences \cite{openai, dpo}. One of the most prominent methods in this domain is Direct Preference Optimization (DPO), which involves fine-tuning a model on a dataset of ranked sample pairs \cite{dpo}. DPO has achieved substantial progress in aligning Large Language Models, owing to its supervised nature and simplicity compared to hyperparameter-heavy reinforcement learning methods like RLHF \cite{openai}. Recently, DPO has been adapted for Diffusion Models, significantly improving overall user preference, prompt adherence, aesthetics, and image structure \cite{diffusion_dpo}.

Diffusion Models are trained to predict pixel-space noise, and pixel space lacks semantic or perceptual structure \cite{lpips}. Latent Diffusion models attempt to bring the diffusion process into the latent space of a Variational Autoencoder (VAE) \cite{sd15}, but this mainly serves to reduce computational costs, and these latent spaces still retain a lot of pixel-level details.

In contrast, it is well-known that the embedding spaces of pretrained deep vision networks are perceptual and highly informative, capable of capturing high-level properties of images \cite{lpips}. This discrepancy motivates us to explore a perceptual objective for aligning diffusion models more effectively with human preferences.

There have been prior attempts to equip diffusion model training objective with some form of perceptual loss \cite{pefl}. These losses are implemented by denoising from noise level $t$ to the initial sample $x_0$, decoding it with VAE, and then feeding it into a corresponding vision network. However, this approach is challenging to optimize effectively.

Notably, recent advancements have shown that the U-Net backbone used in diffusion models — a deep pretrained vision model itself — possesses an embedding space with perceptual properties \cite{ddae, vpd, transferable}. The paper "Diffusion Model with Perceptual Loss" \cite{perceptual_diffusion} proposes utilizing this embedding space for pretraining, showing slightly improved results compared to training without classifier free guidance \cite{cfg}. Building on this, we propose a noise-conditioned perceptual loss for preference optimization. Specifically, we utilize the pretrained encoder of a denoising U-Net, which operates within a noise-conditioned embedding space. This approach allows us to overcome the limitations of pixel-space optimization and directly align with human perceptual features, significantly accelerating the training process.

Additionally, existing image preference datasets are often much larger than necessary and are challenging to train on consumer devices. For example, while Large Language Models benefit from megabytes of preference data, the most popular image preference dataset, Pick-a-Pic, weighs 330GB and contains million of pairs, yet yields less significant results compared to tuning LLMs. Therefore, we analyze and refine the Pick-a-Pic dataset to enable better model training on consumer-grade hardware.

Our contributions are as follows:

\begin{enumerate}
    \item \textbf{Noise-Conditioned Perceptual Preference Optimization (NCPPO)}: We introduce NCPPO, a method that utilizes the U-Net encoder's embedding space for preference optimization. This approach aligns the optimization process with human perceptual features, rather than the less informative pixel space. It can be seamlessly combined with Direct Preference Optimization (DPO), Contrastive Preference Optimization (CPO), and Supervised Fine-Tuning (SFT), further enhancing their effectiveness.

    \item \textbf{Enhanced Training Efficiency}: Our method reduces the compute and training time required for preference optimization. For example, NCPPO achieves superior results with 40\% less computational time.

    \item \textbf{Data Enhancement}: We demonstrate that the large sizes of current datasets, such as the hundreds of gigabytes required for the Pick-a-Pic dataset \cite{pickapic}, are unnecessary. By reducing the size of the Pick-a-Pic dataset by tenfold, we achieve significantly better results.

\end{enumerate}

Embedding the preference optimization process within a noise-conditioned perceptual space provides a more natural and efficient method for aligning diffusion models with human preferences. Our results indicate that this approach not only improves the quality of generated images but also significantly reduces computational costs, making it a promising direction for future research and practical applications.

\section{Related Works}

\subsection{Diffusion Model Fine-Tuning} Various approaches have been developed to fine-tune diffusion models for better alignment with specific objectives, such as conditioning or user preferences. One class of methods uses gradients of explicit reward functions. Classifier Guidance \cite{classifier_guidance, score_based} proposes an inference-time technique to guide generations toward desirable regions using the gradients of a classifier. However, this method requires a noise-conditioned classifier network \cite{cfg}, which makes it impractical in many settings. Unified Guidance \cite{universal_guidance, elucd} instead uses intermediate denoised predictions as inputs for an off-the-shelf classifier. However, these predictions are often out-of-distribution for most vision models due to nonlinearity in sampling trajectories.

Training-time techniques like DRaFT \cite{draft}, ReFL \cite{imagereward}, and AlignProp \cite{alignprop} optimize for maximum differentiable reward by backpropagating gradients through the entire sampling process, which demands substantial GPU memory.

Similar to RLHF in language models \cite{openai}, several works employ Reinforcement Learning (RL) for reward optimization in diffusion models \cite{ddpo, dpok}. These methods frame the generation process as a Markov Decision Process (MDP) with the noise predictor network acting as an agent, using variants of Policy Gradient \cite{ppo} with KL-regularized rewards \cite{rlhf}. While these approaches do not require differentiable rewards, they are often unstable and susceptible to reward hacking.

\begin{figure*}
    \centering
    \includegraphics[width=0.85\linewidth]{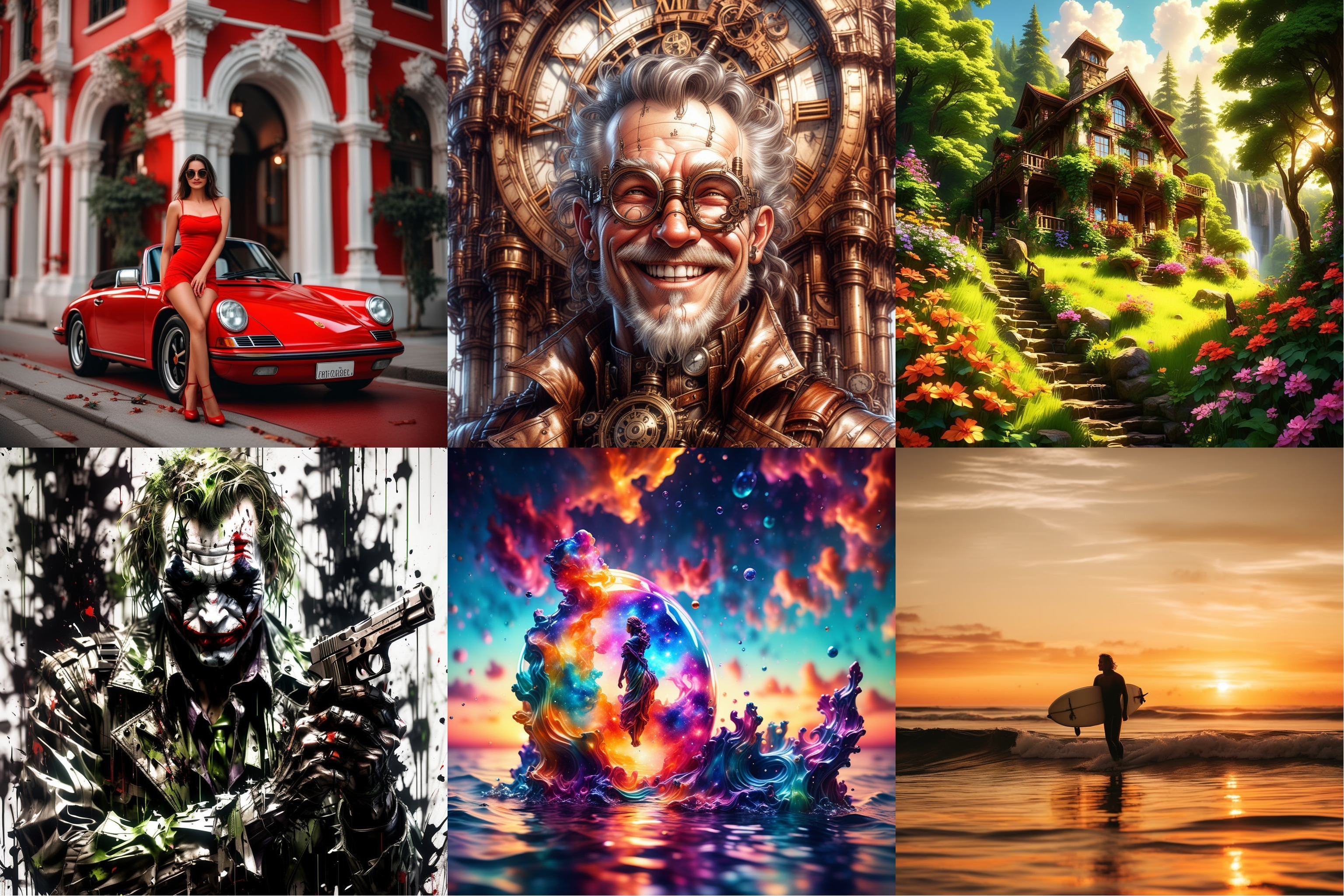}
    \caption{Our method adapts much better to human preferences compared to baseline latent / pixel implementations and can produce extremely high visual appeal alignment.}
    \label{fig:enter-label}
\end{figure*}

\subsection{Direct Preference Optimization} Initially proposed for language models, Direct Preference Optimization (DPO) avoids the instability and complexity of RL methods by reformulating the RLHF objective in terms of an implicit optimal reward \cite{dpo}. Diffusion-DPO \cite{diffusion_dpo} extends this approach to diffusion models, employing conditional score matching objectives for efficient training. Significant efforts have been made to improve the DPO procedure, particularly in the context of large language models (LLMs). These efforts include identifying sources of preference data \cite{rso, leverage_suboptimal}, employing listwise instead of pairwise losses \cite{lipo, pro}, and experimenting with the reference model training, removing, or replacing it \cite{cpo, orpo, tr_dpo}. Despite these advancements in NLP, relatively little has been proposed specifically for text-to-image diffusion models in this area.

\subsection{Diffusion U-Net Properties} Pretrained deep image networks, including U-Nets, have been shown to possess rich, transferable latent spaces that are applicable to various downstream tasks \cite{lpips}. Recent studies indicate that diffusion model backbones, such as U-Net \cite{unet} and DiT \cite{dit}, are no exception. U-Net backbones trained with a diffusion objective can serve as excellent initializations for fine-tuning on downstream image-to-image tasks, such as semantic segmentation or depth estimation \cite{vpd, transferable}, and can even be used for classification with minimal additional training \cite{ddae}. In fact, noise-conditioned classifiers for Classifier Guidance were initialized with U-Net in the original work \cite{classifier_guidance}. Moreover, features from U-Net cross-attention layers have been extensively manipulated to achieve controlled generation and editing \cite{prompt_to_prompt, self_guidance, already_semantic}.

Building on these insights, our work proposes utilizing the U-Net encoder's embedding space for preference optimization in diffusion models, offering a more efficient and perceptually aligned training process.

\section{Preliminaries}

\textbf{Diffusion models} are latent-variable generative models that generate data by iteratively denoising a sample from Gaussian noise \cite{ddpm}. The core idea is to model the data distribution with a two-way Markov chain of gradually noised/denoised latent variables. The diffusion model formulation consists of a fixed forward process, which takes a sample from the data distribution and progressively corrupts it with Gaussian noise, and a parametric reverse process that learns to revert this corruption and effectively recover samples from the data distribution.

In the forward process, a data point $x_0$ is transformed into a noisy version $x_t$ over $T$ discrete timesteps. At each timestep $t$, noise is incrementally added according to a predefined variance schedule $\{\alpha_t\}_0^T$. In terms of samples, this process can be formulated as follows:

\begin{equation}
  \label{eq:forward}
  x_t = \sqrt{\bar{\alpha}_t} x_0 + \sqrt{1 - \bar{\alpha}_t} \epsilon, \quad \epsilon \sim \mathcal{N}(0, \mathbf{I})
\end{equation}

The reverse process involves learning a model $\epsilon_\theta$ that predicts the noise added at each step, directly recovering an estimate of $x_0$. Transition to previous step sample $x_{t-1}$ is defined by the DDPM reverse process as follows:

\begin{equation}
  \label{eq:reverse}
        x_{t-1} = \frac{1}{\sqrt{\alpha_t}} \Big( x_t - \frac{1 - \alpha_t}{\sqrt{1 - \bar{\alpha}_t}} \epsilon_{\theta}(x_t, t) \Big) + \sigma_t \mathbf{z}, \mathbf{z} \sim \mathcal{N}(0, \mathbf{I})
\end{equation}

Diffusion denoising model $\epsilon_\theta$ can be efficiently trained with SGD by minimizing the squared error between the true noise $\epsilon$ and predicted noise $\epsilon_\theta(x_t, t)$. Formally, this loss is defined as follows:

\begin{equation}
  \label{eq:diffusion_loss}
  L_\theta = ||{\epsilon - \epsilon_\theta (x_t, t)}||_2^2, \quad \mathcal{L}(\theta) = \mathbb{E}_{x \sim \mathcal{D}} \left[ L_\theta \right]
\end{equation}

After the denoising model is trained, sampling does not need to involve all $T$ steps and can utilize the SDE or score-based formulation of diffusion models \cite{score_based, elucidating}.

\subsection{Preference Optimization}

Next, we review Direct Preference Optimization (DPO) and Contrastive Preference Optimization (CPO) approaches for diffusion models, establishing the foundation for our proposed Noise-Conditioned Perceptual Preference Optimization (NCPPO).

The task of preference optimization is to fine-tune a generative model such that it produces samples which are more aligned to what humans find preferable. It is assumed that humans express preference according to a latent reward function $r^*$, and that there is a dataset $\mathcal{D} = \{(\mathbf{c}, x^w, x^l)\}$, where $\mathbf{c}$ is a condition, and $x^w$ and $x^l$ are the preferred (winner) and dispreferred (loser) generated samples, respectively.

RLHF \cite{rlhf} first aims to obtain an explicit parameterized reward function $r_\phi$ by fitting Bradley-Terry model on the dataset $\mathcal{D}$ with maximum likelihood.

\begin{figure*}
    \centering
    \includegraphics[width=0.8\linewidth]{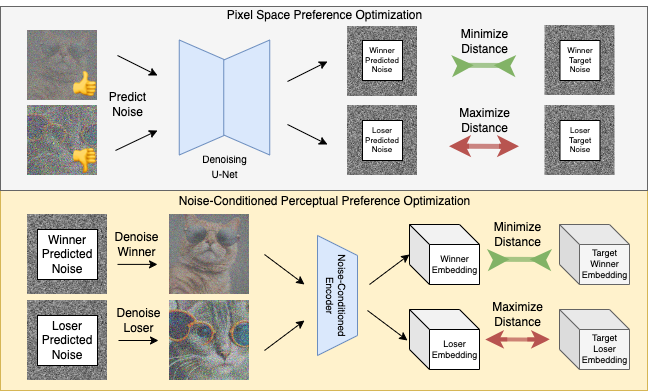}
    \caption{Overall NCPPO pipeline. We optimize preferences inside a Noise-Conditioned embedding
space.}
    \label{fig:pipeline}
\end{figure*}

\begin{equation}
  \label{eq:bt}
  \begin{aligned} 
  &\mathcal{L}_r\left(r_\phi, \mathcal{D}\right)= \\&-\mathbb{E}_{\left(\mathbf{c}, x^w, x^l\right) \sim \mathcal{D}}\left[\log \sigma\left(r_\phi\left(\mathbf{c}, x^w\right)-r_\phi\left(\mathbf{c}, x^l\right)\right)\right]
  \end{aligned}
\end{equation}

With the fitted reward function, it is possible to use Reinforcement Learning to optimize the generative model $p_\theta(x|\mathbf{c})$ as a policy, and employ a form of Policy Gradient to optimize the model. Following \cite{openai}, reward is also modified with regularization by adding KL-divergence term:

\begin{equation}
  \max_{p_\theta} \mathbb{E}_{\mathbf{c} \sim \mathcal{D}_c, x \sim p_\theta\left(x | \mathbf{c}\right)} {\left[r\left(\mathbf{c}, x\right)\right] } -\beta\mathbb{D}_{\mathrm{KL}}\left[p_\theta\left(x | \mathbf{c}\right) \| p_{\mathrm{ref}}\left(x | \mathbf{c}\right)\right]
\end{equation}

DPO elegantly restates $L_{RLHF}$ without the need for RL Policy Gradient estimators. First, it considers the solution to the given optimization problem, optimal policy $p^*(x|\mathbf{c})$ under reward function $r$, and rearranges it into an expression in terms of optimal policy:

\begin{equation}
  p_\theta^*\left(x | \mathbf{c}\right) = \frac{p_{reference}\left(x | \mathbf{c}\right) \exp \left(r\left(\mathbf{c}, x\right) / \beta\right)}{\sum_{x} p_{\mathrm{ref}}\left(x | \mathbf{c}\right) \exp \left(r\left(\mathbf{c}, x\right) / \beta\right)}
\end{equation}

Plugging it into the Bradley-Terry model from equation \eqref{eq:bt}, we arrive at the DPO objective:

\begin{equation}
    \begin{aligned}
    &\mathcal{L}_{\mathrm{DPO}}(\theta) =\\
    &-\mathbb{E}_{\left(\mathbf{c}, x^w, x^l\right) \sim \mathcal{D}} \left[\log \sigma \left( \beta \log \frac{p_\theta(x^w | \mathbf{c})}{p_{ref}(x^w | \mathbf{c})} - \beta \log \frac{p_\theta(x^l | \mathbf{c})}{p_{ref}(x^l | \mathbf{c})} \right) \right]
    \end{aligned}
\end{equation}

Through several approximations and careful rearrangement of expectations using Jensen's inequality, Wallace et al. \cite{diffusion_dpo} find a way to rewrite the DPO objective for diffusion models in the following form:

\begin{equation}
  \label{eq:dpo_loss}
  \begin{aligned}
  &\mathcal{L}_{\text{DiffusionDPO}}(\theta)=\\ &-\mathbb{E}_{\left(\mathbf{c}, x^w, x^l\right) \sim \mathcal{D}} \left[\log \sigma\left(-\beta T \left( (L_\theta^w - L_{ref}^w) - (L_\theta^l - L_{ref}^l) \right)\right) \right]
  \end{aligned}
\end{equation}

where $L$ corresponds to the diffusion loss from \eqref{eq:diffusion_loss}, calculated over winners, losers, current and reference models, respectively.

Contrastive Policy Optimization (CPO) is derived from the DPO objective by substituting the reference policy  $p_{ref}(x|\mathbf{c})$ with the optimal policy $p_{w}(x|\mathbf{c})$, defined such that  $p_{w}(x^w|\mathbf{c})$ = 1 and \\ 0 $\leq$ $p_{w}(x^l|\mathbf{c}) $ $\leq$  1. By incorporating this substitution into the original DPO objective, Xu et al. \cite{cpo} derive the following formulation:

\begin{equation}
\begin{aligned}
      &\mathcal{L}_{\text{CPO}}(\theta) = \\ &-\mathbb{E}_{\left(\mathbf{c}, x^w, x^l\right) \sim \mathcal{D}} \Big{[}\log \sigma ( \beta \log p_\theta(x^w | \mathbf{c}) - \\ &\beta \log p_\theta(x^l | \mathbf{c}) + \beta \log p_{ref}(x^l | \mathbf{c}) ) \Big{]}
\end{aligned}
\end{equation}

After performing a series of intermediate calculations, the CPO objective simplifies to:

\begin{equation}
  \begin{aligned}
  &\mathcal{L}_{\text{CPO}}(\theta) = \mathbb{E}_{(\mathbf{c}, x^w, x^l) \sim \mathcal{D}} 
  \left(\mathbf{c}, x^w, x^l
  \right)
  \sim 
    \\ &
    \mathcal{D}
  \left[
  \log \sigma 
  \left( \beta \log p_\theta(x^w | \mathbf{c})
  \right)
   -
  \beta \log p_\theta
  \left(
  x^l 
  | \mathbf{c}
  \right) 
  \right]
  \\ &
  -  \lambda \mathbb{E}_{(\mathbf{c},x^w) 
  \sim \mathcal{D}} 
  \left[ 
  \beta \log p_\theta(x^w | \mathbf{c}) \right]
  \end{aligned}
\end{equation}

where the second term provides additional guidance for the policy.
Applying this formulation to the diffusion problem results in the final objective:

\begin{equation}
    \begin{aligned}
      &\mathcal{L}_{\mathrm{DiffCPO}}(\theta)= 
      \\ & -
      \mathbb{E}_{\left(\mathbf{c}, x^w, x^l\right) \sim \mathcal{D}} \left[\log \sigma\left(-\beta T \left( L_\theta^w  
      -
      L_{\theta^l}
      \right )\right) \right] +
      \lambda \mathbb{E}_{(\mathbf{c},x^w) \sim \mathcal{D}} \left[ L_\theta^w \right]
    \end{aligned}
\end{equation}

\section{Noise-Conditioned Preference Optimization}

Inspired by the properties of human perception and the successful alignment of large language models (LLMs) within informative embedding spaces, our goal is to perform diffusion preference optimization within a similarly informative perceptual embedding space. As the preference optimization objectives in Equation \eqref{eq:dpo_loss} replace logits with a diffusion squared error objective, we first introduce a diffusion perceptual loss, similar to the approach in \cite{perceptual_diffusion}.

Currently, there are no pretrained open-source noise-conditioned encoder networks that operate in the SD1.5 and SDXL VAE latent space, except for the encoder of a pretrained U-Net. A naive way of using an off-the-shelf network like CLIP would require predicting $x_0$ directly from arbitrary timestep, and further decoding it with VAE decoder, introducing distribution shifts along the way. Meanwhile, U-Net is already noise-conditioned, operates in latent space, incorporates text condition, and has been shown to exhibit the same perceptual properties as other pre-trained vision networks \cite{ddae}.

Let $f(x_{t'}, \mathbf{c}, t')$ denote the downsampling stack of pre-trained U-Net, evaluated as some timestep $t'$ - this is our perceptual encoder. We choose to evaluate it at $t'=t-1$, previous timestep from the one noise prediction was obtained from, recovered during training by performing DDPM reverse step \eqref{eq:reverse}. To obtain ground truth embedding, we perform reverse step on true noise.
We subscript $x_{t'}$ with indications of which noise is used to perform the step: for example, obtaining a sample for winner and optimized model looks like this:

\begin{equation}
    \begin{aligned}
      & x_{\theta, t'}^w = x_{\theta, t-1}^w = \\ & \frac{1}{\sqrt{\alpha_t}} \Big( x_t^w - \frac{1 - \alpha_t}{\sqrt{1 - \bar{\alpha}_t}} \epsilon_{\theta}(x_t^w, t, \mathbf{c}) \Big) + \sigma_t \mathbf{z}, \text{ } \mathbf{z} \sim \mathcal{N}(0, \mathbf{I})
    \end{aligned}
\end{equation}


This results in the following perceptual diffusion objective:

\begin{equation}
  PL_\theta = ||{f(x_{\theta, t'}, \mathbf{c}, t') - f(x_{t'}, \mathbf{c}, t')}||_2^2
\end{equation}

We use this formulation in place of the standard diffusion loss in Preference Optimization objectives. Denote the losses for the winner, loser, winner (reference), and loser (reference) as follows:

\begin{align}
   &PL_\theta^w = ||{f(x_{\theta, t'}^w, \mathbf{c}, t') - f(x_{t'}^w, \mathbf{c}, t')}||_2^2, \\
   &PL_{\text{ref}}^w = ||{f(x_{\text{ref}, t'}^w, \mathbf{c}, t') - f(x_{t'}^w, \mathbf{c}, t')}||_2^2 \\
  &PL_\theta^l = ||{f(x_{\theta, t'}^l, \mathbf{c}, t') - f(x_{t'}^l, \mathbf{c}, t')}||_2^2, \\
  &PL_{\text{ref}}^w = ||{f(x_{\text{ref}, t'}^l, \mathbf{c}, t') - f(x_{t'}^l, \mathbf{c}, t')}||_2^2
\end{align}

\textbf{DPO target:} We use the embeddings in the DiffusionDPO loss, simply replacing the noise terms with the embeddings obtained from the encoder:

\begin{equation}
  \begin{aligned}
  &\mathcal{L}_{\mathrm{DiffusionDPO}}(\theta)= \\ &-\mathbb{E}_{\left(\mathbf{c}, x^w, x^l\right) \sim \mathcal{D}} \left[\log \sigma\left(-\beta T \left( (PL_\theta^w - PL_{ref}^w) - (PL_\theta^l - PL_{ref}^l) \right)\right) \right]
  \end{aligned}
\end{equation}


\textbf{Contrastive Preference Optimization target:} Following the original paper and replacing the noise term with perceptual embeddings, we get:

\begin{equation}
\begin{aligned}
  & \mathcal{L}_{\text{DiffusionCPO}}(\theta)= \\ &-\mathbb{E}_{\left(\mathbf{c}, x^w, x^l\right) \sim \mathcal{D}} \left[\log \sigma\left(-\beta T \left( PL_\theta^w - PL_\theta^l \right)\right) \right] + \\ & \lambda \mathbb{E}_{(\mathbf{c},x^w) \sim \mathcal{D}} \left[ PL_\theta^w \right]
\end{aligned}
\end{equation}

In the case of CPO the coefficient $\beta$ is different since we omit the reference model.


\section{Enhancing Preference Data}

Now, we want to highlight our second contribution, which demonstrates how image preference datasets should be designed. We find that with more carefully curated data, our NCP-DPO method significantly outperforms the baseline DPO even further while the overall quality of both models (DPO, NCP-DPO) increases significantly. To illustrate this, we provide synthetic win rates based on HPSv2 and PickScore, comparing NCP-DPO on both the filtered and original Pick-a-Picv2 datasets. The primary issue with the original dataset lies in the contradictory examples it contains a significant portion of the images serve as both winners and losers in different pairs. For instance, suppose we have three images for a single prompt, $x_1 < x_2 < x_3$ where $<$ denotes preference. Now, let’s recall the gradient for the Bradley-Terry model:

\begin{equation}
\begin{aligned}
    &\nabla_\theta \mathcal{L}_{\text{DPO}}(x^w, x^l) =\\& -\beta \mathbb{E}_{(c, x^w, x^l) \sim \mathcal{D}} \Big{[} \sigma(r_\theta(x^l) - r_\theta(x^w)) \times \\& (\nabla_\theta \log p(x^w) - \nabla_\theta \log p(x^l)) \Big{]}
\end{aligned}
\end{equation}

Now let's write down the sum of two gradients computed on $x_1, x_2$ and $x_2, x_3$ pairs.

\begin{equation}
\begin{aligned}
&\nabla_\theta \mathcal{L}_{\mathrm{DPO}}(x_2, x_1) + \nabla_\theta \mathcal{L}_{\mathrm{DPO}}(x_3, x_2)= \\ &  -\beta \left[ \sigma(r_\theta(x_1) - r_\theta(x_2)) \cdot (\nabla_\theta \log p(x_2) - \nabla_\theta \log p(x_1)) \right] 
\\ &- \beta \left[ \sigma(r_\theta(x_2) - r_\theta(x_3)) \cdot (\nabla_\theta \log p(x_3) - \nabla_\theta \log p(x_2)) \right]
\end{aligned}
\end{equation}

Notice how terms $\beta \sigma(r_\theta(x_1) - r_\theta(x_2)) \cdot \nabla_\theta \log p(x_2)$ and $\beta \sigma(r_\theta(x_2) - r_\theta(x_3)) \cdot \nabla_\theta \log p(x_2)$ partially cancel each other out, adding noise instead of a clear signal. When excluding contradictory examples, the training process becomes more stable and predictable. Synthetic comparisons for original and filtered dataset are shown in Table \ref{tab:winrate_new}.

\begin{table}[]
    \centering
    \begin{tabular}{lcc}
        \hline
        NCP-DPO vs DPO& PickScore Winrate & HPSv2 Winrate \\
        \hline
        SDXL & 54.3  & 55.7 \\
        SDXL (filtered data) & \textbf{65.8} & \textbf{67.1} \\
        SD1.5 & 56.8 &  59.1 \\
        SD1.5 (filtered data) & \textbf{64.6} & \textbf{71.3} \\
        \hline
    \end{tabular}
    \caption{Synthetic winrates for NCP-DPO models against DPO. Filtering data improves both NCP-DPO and DPO and makes our method even more powerful. }
    \label{tab:winrate_new}
\end{table}

\begin{figure*}
    \centering
    \includegraphics[width=1\linewidth]{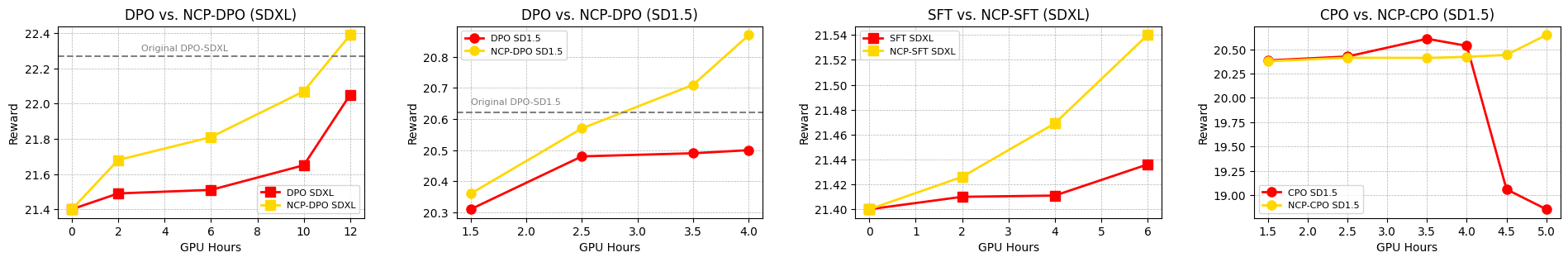}
    \caption{We evaluate training speed using PickScore on a Pick-a-Pic validation set. Our method significantly accelerates the learning process compared to baseline methods like DPO and Supervised Fine-Tuning while also achieving superior quality. Contrastive Preference Optimization is very unstable due to the lack of a reference model, but we demonstrate that our method provides a regularization effect as well.}
    \label{fig:speeds}
\end{figure*}

To further analyze the impact of contradictory examples, we created three additional mini versions of the Pick-a-Pic dataset. For each prompt, we selected their absolute winner $x$ (images that have 0 losses or draws and more than 1 win), their latest losers $y$, and images $z$ that are the latest losers for $y$. Results from Table \ref{tab:winrate} show an intuitive fact that winners must be as good as possible and contradictory samples decreases the overall quality.

\begin{figure*}[h]
    \centering
    \includegraphics[width=0.9\linewidth]{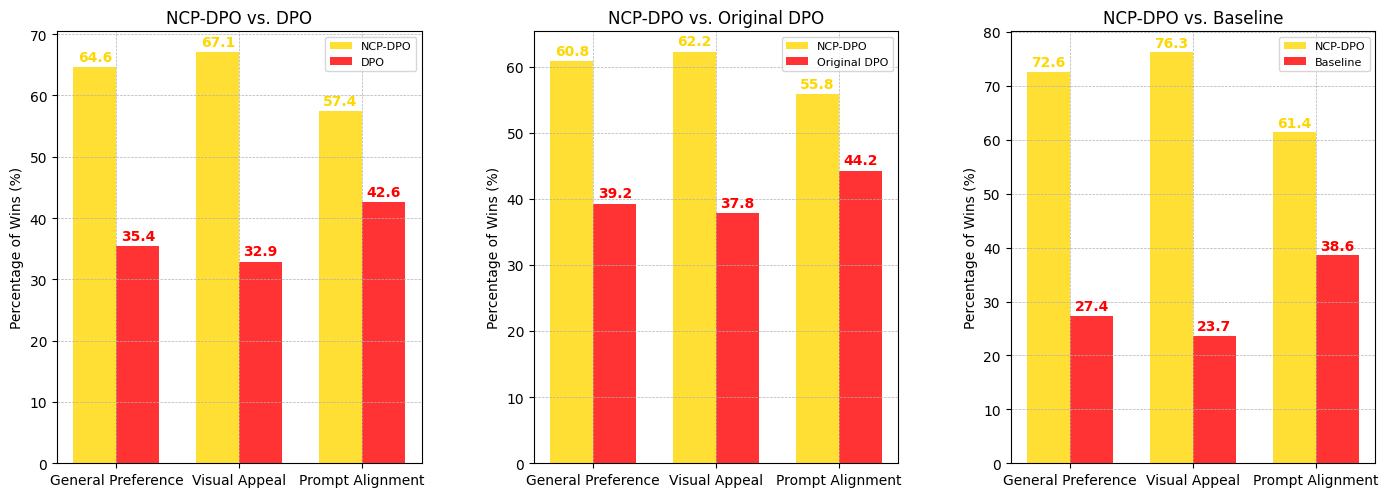}
    \caption{Side-by-side real human preferences comparison for different SDXL models using PartiPrompts benchmark. We compare
NCP-DPO with 1) Our own DPO-SDXL 2) Original published DPO-SDXL 3) Baseline model with
no preference optimization. Our method significantly improves Direct Preference Optimization. All
models are trained on the same data.}
    \label{fig:sbs}
\end{figure*}

\begin{table}[ht]
    \centering
    \begin{tabular}{lcc}
        \hline
        Comparison & PickScore Median \\
        \hline
        $x>y$ & 21.98 \\
        $x > y \cup x>z$ & 22.05 \\
        $x>y \cup y>z$&  21.96\\
        \hline
    \end{tabular}
    \caption{Synthetic PickScore rewards for different dataset setups. $x>y$ stays for mini dataset with absolute winners across all prompts and their latest losers, $x > y \cup y > z$ denotes the same dataset with added images $z$ that are losers to $y$, $x>y \cup x>z$ denotes the same as previous but instead the winners are always absolute. We train these models for 250 steps keeping the same hyperparameters.}
    \label{tab:winrate}
\end{table}

\section{Experiments}


In this section we empirically validate our proposed contributions. As we have two independent contributions (perceptual objective for preferences, data improvement), our goal is to show that each one of them significantly improves the overall quality and training speed in terms of synthetic reward independently. We observe that combining perceptual objective with enhanced data improves the model even further compared to original DPO trained on enhanced data. 

\subsection{Setup} We fine-tune Stable Diffusion v1.5 (SD1.5) \cite{sd15} and Stable Diffusion XL (SDXL) \cite{sdxl} models.
We run our largest experiments in terms of compute on a Pick-a-Pic v2 dataset \cite{pickapic}, following a setup from \cite{diffusion_dpo} that consists of 851,293 pairs, with 58,960 unique prompts, obtained from versions of SDXL and DreamLike (a fine-tune of SD1.5). We train both original DPO and NCP-DPO to make the comparison fair, however, we also compare our NCP-DPO with original published SDXL model. This experiment is validated only via side-by-side user study.
To enhance the dataset, we leave only absolute winners across all prompts and get a version with 87,687 pairs and 53,701 unique prompts. Here, we score a uniform subset of checkpoints during training to show that Noise-Conditioned Perception works better and quicker adapts to human preferences. Both dataset and best model checkpoint based on reward will be publicly available. 
All our checkpoints, including baselines, are obtained with rank 64 LoRA \cite{lora} for U-Net weights, contrary to full-finetunes from \cite{diffusion_dpo, diffusion_kto}. We also use 8-bit AdamW optimizer \cite{adamw} with default hyperparameters and half precision for non-trainable weights everywhere.
For large experiments we use hyperparameters exactly like in the original Diffusion-DPO paper. We increase the learning rate for smaller experiments in 3 times. All experiments were done using 4x Nvidia H100 GPU and Intel(R) Xeon(R) Platinum 8480+ CPU processor.

\subsection{Cheaper and Quicker} To demonstrate the efficiency of our proposed method, we use synthetic rewards \cite{hpsv2, pickapic} to evaluate the generated images. HPSv2 is considered the most reliable, as it has been shown to best correspond with human judgment. We generated samples for the validation set of the Pick-a-Pic-v2 dataset, which consists of 500 unique prompts, following the exact Diffusion-DPO setup. For each method—Supervised Fine-Tuning, DPO, CPO, and their respective improvements, NCP-SFT, NCP-DPO, and NCP-CPO—we generated images with the same amount of training time. Training speeds are illustrated in Fig. \ref{fig:speeds}. We observed that NCP-DPO-SD1.5 significantly outperformed the original DPO-SD1.5 in terms of PickScore reward, achieving this with only 3.5 GPU hours of training. A similar trend was seen with SDXL: our approach not only outperformed the original SDXL-DPO when trained on the same data but also produced a superior model in just 12 GPU hours when trained on the improved Pick-a-Pic dataset. Additionally, Contrastive Preference Optimization lacks a reference model to stabilize the training process, making the model more prone to overfitting and divergence. However, NCP-CPO effectively mitigates this issue by matching the trainable U-Net's embeddings with a frozen copy, providing significant regularization.

\subsection{Side-by-Side} To validate the method overall improvements, we follow original Diffusion-DPO paper human feedback annotation setup, comparing generations for PartiPrompts \cite{parti} under three different criteria: Q1 General Preference (Which image do you prefer given the prompt?), Q2
Visual Appeal (prompt not considered) (Which image is
more visually appealing?) Q3 Prompt Alignment (Which
image better fits the text description?). Each generation had 3 independent annotations for each criteria. Image is considered a winner according to a majority votes.

We validate only DPO method with the same dataset as in original paper. Our human annotations show superior results. We compare our largest run with the original  DPO-SDXL published weights and a baseline method DPO trained by us with exactly the same hyperparameters as in NCP-DPO version for a fair comparison.  Results are shown on Fig.\ref{fig:sbs}.

\section{Conclusion}
Optimizing Diffusion models for human preferences using Noise-Conditioning Perception is a natural and powerful way to guide the training process of diffusion models. In this work, we have shown that using U-Net's own encoder embedding space is already a strong baseline for improving both the speed of adaptation to human preferences and the overall quality of the model compared to original methods.

\section{Limitations \& Future Work} While Noise-Conditioned Perception works well for diffusion models based on the U-Net architecture, it remains a question for future work whether this approach works that well with Diffusion Transformer architecture \cite{dit}. We leave this investigation to future research. 

\section*{Acknowledgment}
The work of I.~Makarov on Section 1 and 2 was supported by a grant, provided by the Ministry of Economic Development of the RF in accordance with the subsidy agreement (agreement identifier 000000C313925P4G0002) and the agreement with the Ivannikov Institute for System Programming of the Russian Academy of Sciences dated June 20, 2025 No. 139-15-2025-011.

\bibliographystyle{unsrtnat}
\bibliography{references}

\end{document}